\documentclass[10pt,twocolumn,letterpaper]{article}

\usepackage{wacv}              

\usepackage{graphicx}
\usepackage{amsmath}
\usepackage{amssymb}
\usepackage{booktabs}

\usepackage{subcaption}
\usepackage{comment}
\usepackage{tabularx}

\usepackage{cite}
\usepackage{amsmath,amssymb,amsfonts}
\usepackage{algorithmic}
\usepackage{textcomp}
\usepackage{xcolor}

\usepackage{amsthm,enumitem}

\usepackage{soul}
\usepackage{multirow}

\usepackage{comment}
\usepackage{enumitem}
\usepackage{booktabs}

\usepackage[pagebackref,breaklinks,colorlinks]{hyperref}

\usepackage[capitalize]{cleveref}
\crefname{section}{Sec.}{Secs.}
\Crefname{section}{Section}{Sections}
\Crefname{table}{Table}{Tables}
\crefname{table}{Tab.}{Tabs.}

\def\wacvPaperID{2197} 
\def\confName{WACV}
\def\confYear{2024}

\begin{document}

\title{Fused Classification For Differential Face Morphing Detection}

\author{Iurii Medvedev\\
\textit{Institute of Systems}\\
\textit{and Robotics,}\\
\textit{University of Coimbra,}\\
Coimbra, Portugal\\
{\tt\small iurii.medvedev@isr.uc.pt}
\and
Joana Alves Pimenta\\
\textit{$^1$Institute of Systems}\\
\textit{and Robotics,}\\
\textit{University of Coimbra,}\\
\textit{Coimbra, Portugal}\\
{\tt\small joana.pimenta@isr.uc.pt}
\and
Nuno Gonçalves $^{1,2}$\\
\textit{$^2$Portuguese Mint and Official}\\
\textit{Printing Office (INCM),}\\
\textit{Lisbon, Portugal}\\
{\tt\small nunogon@deec.uc.pt}
}
\maketitle

\begin{abstract}
Face morphing, a sophisticated presentation attack technique, poses significant security risks to face recognition systems. Traditional methods struggle to detect morphing attacks, which involve blending multiple face images to create a synthetic image that can match different individuals. In this paper, we focus on the differential detection of face morphing and propose an extended approach based on fused classification method for no-reference scenario. We introduce a public face morphing detection benchmark for the differential scenario and utilize a specific data mining technique to enhance the performance of our approach. Experimental results demonstrate the effectiveness of our method in detecting morphing attacks.
\end{abstract}

\section{Introduction}
The development of deep learning techniques in recent years has led to significant progress in the field of face recognition, but sophisticated presentation attack techniques, such as face morphing, continue to pose security risks that require new protection solutions. Face morphing involves merging/blending of two or more digital face images to create a synthetic image that can share the biometric properties of original images and match different individuals. Such generated images it can be difficult to detect using traditional human or computer-based face recognition methods. 

The risks associated with face morphing are not hypothetical; they have been demonstrated through real-world incidents. One notable example occurred in 2018 when a German activist exploited face morphing techniques to issue an authentic German passport using a morphed face image of Federica Mogherini (at that time High Representative of the Union for Foreign Affairs and Security Policy) blended with their own photo \cite{morphing_spiegel}. This incident highlighted the potential for face morphing attacks to deceive identity verification systems and emphasizes the need for effective detection methods. Additionally, face morphs have been occasionally detected during border control procedures, raising concerns about the circulation of morphed documents. Some recent investigations \cite{torkar2023morphing} acknowledged the presence of morphing cases, indicating the tangible risks and uncertainties surrounding the prevalence of such documents. These real-world examples underscore the urgency of robust morphing detection approaches to mitigate the security risks associated with face morphing attacks.
That is why face morphing and its detection methods have gained interest in both industry \cite{iMARS_project} and academia \cite{nist_conference_2022}.

Morphing detection methods in facial biometric systems can be categorized into two  pipelines based on the processing scenario. The \textit{no-reference} morphing attack detection algorithm is designed to detect morphing in a single image, with a focus on mitigating the risks of accepting manipulated images during the \textit{enrollment} process, where successful acceptance of forged images can lead to the issuance of an authentic document that could deceive the face recognition system. 

On the other hand, the \textit{differential} morphing attack detection algorithms involve acquiring live data from an authentication system to provide reference information for detecting morphing attacks. This usually occurs during \textit{automatic border control} and such approaches aim to identify discrepancies between the presented face and the stored biometric data, enabling the system to detect potential morphing attempts in real-time and prevent unauthorized access with malicious ID documents (documents with accepted face morphs).


In this paper, we focus on differential face morphing detection and propose a novel deep learning method that incorporates sophisticated face recognition tasks and employs a fused classification scheme for morphs classification. We follow the no-reference MorDeephy\cite{MorDeephy} approach and adopt it methodology and data for the differential case.


Additionally, we extend benchmark utilities, which are proposed in \cite{MorDeephy} with a public face morphing detection benchmark for differential scenario.

\section{Related Work}

\subsection{Face Recognition}
Contemporary face recognition methods heavily rely on the utilization of deep learning techniques, particularly convolutional neural networks (CNNs), which have proven to be highly effective in extracting discriminative features from unconstrained facial images. \cite{ImageNet_cite}. These networks possess the ability to learn complex patterns and structures, making them well-suited for tackling the challenges associated with facial pattern recognition tasks.

Various deep learning strategies are employed for face recognition, all aimed at extracting low-dimensional facial representations - deep face features with high discriminatory capabilities. 

For example, metric learning techniques focus on explicit optimizing the face representation by contrasting pairs of matched and non-matched samples with similarity metric \cite{facenet}. Achieving reliable convergence with these methods necessitates extensive datasets and advanced sample mining techniques.

Classification-based methods have received major attention 
and they are are better represented 
in recent academic research. These methods focus on learning face representation implicitly through a closed-set identity classification task \cite{deepid2_plus_paper}. Deep networks in these approaches encapsulate face representation in the last hidden layer and typically employ various softmax-based loss functions \cite{deepid2_plus_paper}.

To achieve better discriminate properties of deep facial features various techniques are used.  For instance, explicit compacting of intra-class features to their center \cite{centerface_paper} or several types of marginal restrictions, which address inter-class discrepancy \cite{arcface_paper,equalized_margin_paper
}.
Many recent works were focused on investigating sample-specific learning strategies, which are driven by various characteristics, such as sample quality \cite{
QualFace2}, hardness of classification  \cite{Huang_2020_CVPR}. 
Some works use on properties of embedding as a proxy for the image quality (like norm of the features) \cite{Magface, AdaFace}, or rely on artificial assignment by known data augmentation \cite{towards_face_recognition}. These approaches try the control the feature distribution in the discriminative feature domain.

\subsection{Face Morphing}
Modern face recognition systems can very accurately match images of individuals, however they are still vulnerable to various malicious presentation attacks. Face morphing allows do design such attack and drastically increase the probability for face recognition network to return matched embeddings for unmatched biometric samples, especially in the cases when the thresholds of face recognition systems are not set to support critically low false match rates. 

Basic landmark based face morphs were first investigated by Ferrara \textit{et al.} \cite {magic_passport}. Face morphing was performed directly in the image spatial domain by the face landmark alignment, image warping and blending. 
Various morphing algorithms mentioned in the literature follow this strategy \cite{ubo_morpher}. 

The field of face morphing has witnessed significant advancements with recent breakthroughs in Deep Learning techniques, leading to the development of several innovative tools and methodologies.
Face morphing has witnessed significant advancements with recent breakthroughs in Deep Learning techniques.
Generative Adversarial Networks (GANs) have emerged as a prominent and widely utilized approach in various generative tasks, including face morphing. 
MorGAN \cite{morGAN} approach pioneered this tool for face morphing generation. The StyleGAN \cite{styleGAN} approach introduced a latent domain representation to control various aspects of the generated image, which enabling to generate high-quality face  morphs without blending artifacts. The MIPGAN \cite{MIPGAN_morphing_paper} method optimized StyleGAN specifically for face morphing, preserving the identity of the generated morphed face image. The diffusion autoencoders for face morphing were proposed by MorDIFF\cite{MorDIFF} to generate smooth and high-fidelity face morphing attacks. 

\subsection{Face Morphing Detection}
Initially, the problem of face morphing detection focused on the no-reference scenario, where validation decisions were based on single image  presentations. However, considering practical concerns, it became valuable to explore a differential approach that simulates the process of document verification by border control officers.

No-reference face morphing detection algorithms initially relied on analyzing local image characteristics like  Binarized Statistical Image Features (BSIF) \cite{detecting_face_morph_1} or sensor noise (Photo Response Non-Uniformity) \cite{PRNU_2}, texture features \cite{Towards_morphing_detection},  local features in frequency and spatial image domain \cite{face_morphing_fd} or fusion of various features \cite{face_rec_vulnerability_morph}.
Deep learning methods for the no-reference case typically involve binary classification of pretrained face recognition features \cite{face_morphing_dnn}, which can be combined with local texture characteristics \cite{face_morphing_using_general_purpose_fr}. 
Additional pixel-wise supervision \cite{PW-MAD} or attention mechanism \cite{AttentionMorphing} can be applied.
MorDeephy method \cite{MorDeephy} generalized single image morphing detection to unseen attacks by additional feature regularisation with face recognition task.

In contrast to no-reference case,  differential face morphing detection is closely correlated with face recognition, as the discriminability of deep face representation usually helps combating attacks in this scenario.

For instance, many differential approaches rely on classification of pretrained  deep features for face recognition \cite{face_morphing_MAD}. 

Borghi et al. \cite{Double_Siamese_Morphing} conducted differential morphing detection through the fine-tuning of pretrained networks within a sophisticated framework involving identity verification and artifacts detection modules.

Qin et al. \cite{FMD_Feature_Wise_Supervision} proposed a method for detecting and locating face morphing attacks for both (single-image and differential) scenarious. The authors utilise feature-wise supervision, which provides better characterization of morphing patterns and localization of morphed areas. 

\begin{figure*}
\begin{center}
  \includegraphics[width=0.99\linewidth]{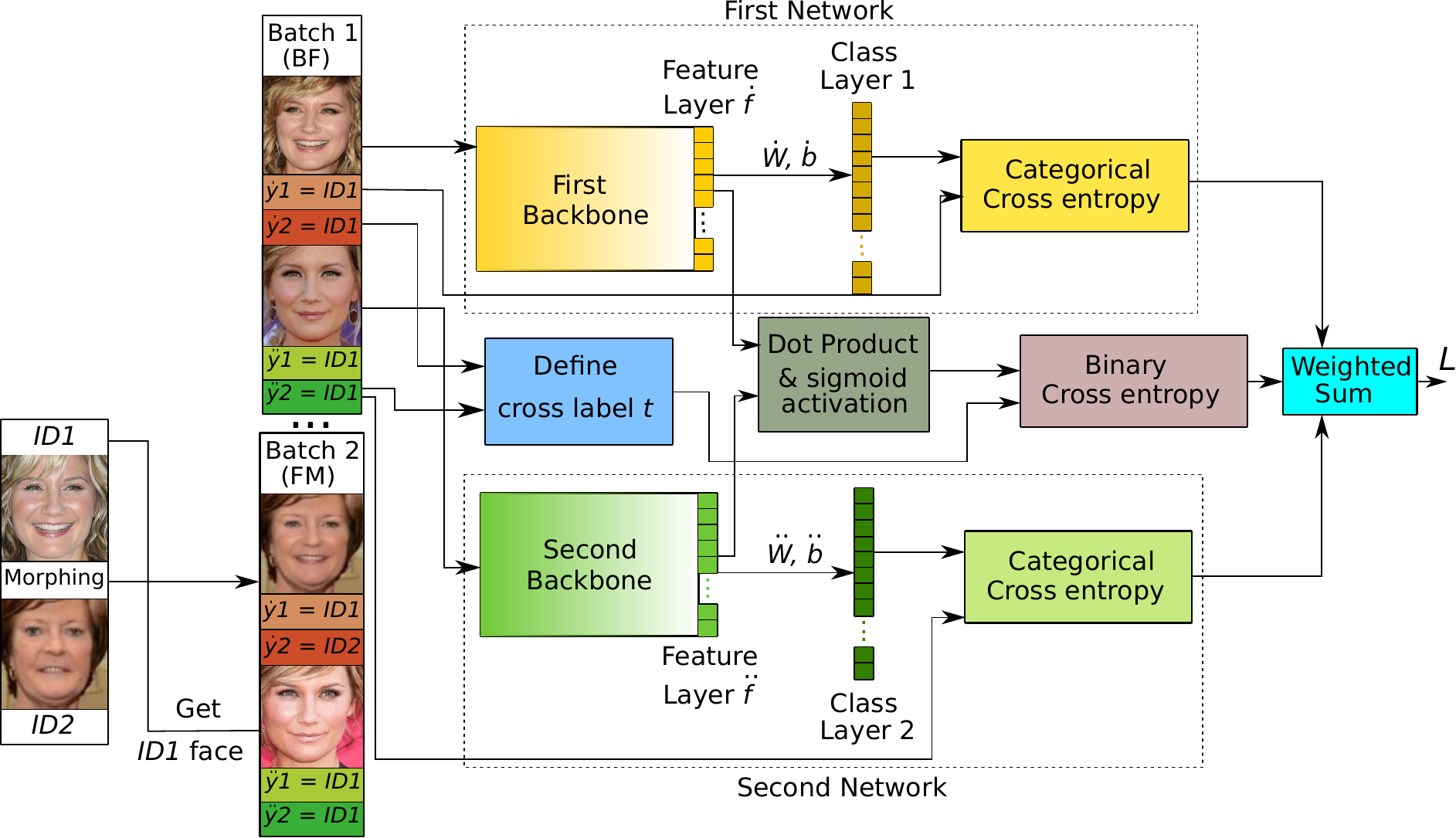}
\end{center}
   \caption{Schematic of the proposed D-MAD method. For simplicity of visualization batch contains a single image pair.}
\label{fig:mdf_schematic}
\end{figure*}

Ferrara et al. \cite{face_demorphing} presented an alternative approach to the differential scenario, which involves reverting morphing by retouching the testing face image using a trusted live capture. This technique aims to unveil the true identity of the legitimate document owner.

In this work we propose to extend the no-reference MorDeephy\cite{MorDeephy} approach for the differential morphing detection scenario by adopting the methodology and data mining techniques to the differential pipeline.

\section{Methodology} 
\label{Methodology}

From the original work \cite{MorDeephy} we inherit the S-MAD methodology, which require several modifications for the differential case. 
Recall that the face morphing detection here is made by the behavior of deep face features, which is achieved by regularizing the morphing detection with face recognition task. The definition of the task is motivated by ubiquity of classifying the face morphs (since they belong to 2 or more identities). This leads to the setup with two separate CNN-based deep networks that treat bona fide samples similarly but handle morphed samples differently. These networks does not share weights and are not trained in a contrastive manner, where positive and negative pairs are matched. Both networks learn high-level features through classification tasks, with each network assigning different identity labels to face morphs. The \textit{First Network} labels them based on the original identity from the first source image, while the \textit{Second Network} labels them according to the second original label.

In comparison to the S-MAD scenario, where the same image is sent to both networks, the D-MAD case imply processing a pair of images.
That is why the fused classification strategy is adopted to the D-MAD in the following way (see Fig. \ref{fig:mdf_schematic}).

We keep the assumption of associating each image with two identity labels $y1$ and $y2$, which are defined basing on the image origin. For the Bona Fide samples those labels are the same (copied from the original face image label), when for the Morphs those labels are different and are taken from the source face images.
The sampling process for the \textit{First Network} is not changed. For instance the image $\dot{I}$ with $\dot{y}1_{\dot{I}}$ and $\dot{y}2_{\dot{I}}$ is sampled for the input. For the input of the \textit{Second Network} the image $\ddot{I}$ (complementary to $\dot{I}$) with $\ddot{y}1_{\ddot{I}}$ and $\ddot{y}2_{\ddot{I}}$ is selected with a condition that $\ddot{y}1_{\ddot{I}} = \dot{y}1_{\dot{I}}$. The loss function components require the following rules. The identity classification for the \textit{First Network} is made by the $\dot{y}1_{\dot{I}}$, and by the $\ddot{y}2_{\ddot{I}}$ for the \textit{Second Network}. The ground truth cross label for the morphing binary classification is made by matching $\dot{y}2_{\dot{I}}$ and $\ddot{y}2_{\ddot{I}}$.

It is important to note that such formulation allow both images $\dot{I}$ and $\ddot{I}$ to be Morphs. However to match the D-MAD scenario, where the \textit{Live Enrollment} image is genuine and trusted, we supervise selecting the $\ddot{I}$ as a Bona Fide sample. 

Due to the above modifications the formulation of the identity classification softmax-based loss components is transformed as follows:

\begin{equation}
\centering
\label{eq:soft_class_1}
    L_{1} = -\frac{1}{N}\sum_{i}^{N} \log (\frac{e^{\dot{W}_{\dot{y}1_i}^{T}\dot{f}_{i}+\dot{b}_{\dot{y}1_i}}}{ \sum_{j}^{C} e^{\dot{f}_{\dot{y}1_j}}}) 
\end{equation}

\begin{equation}
\centering
\label{eq:soft_class_2}
    L_{2} = -\frac{1}{N}\sum_{i}^{N} \log (\frac{e^{\ddot{W}_{\ddot{y}2_i}^{T}\ddot{f}_{i}+\ddot{b}_{\ddot{y}2_i}}}{ \sum_{j}^{C} e^{\ddot{f}_{\ddot{y}2_j}}}),
\end{equation}
where $\left \{  \dot{f}_i, \ddot{f}_i \right \}$ denote the deep features of the $i-th$ sample pair,   $\left \{  \dot{W}, \ddot{W} \right \}$ and $\left \{  \dot{b}, \ddot{b} \right \}$ are weights and biases of last fully connected layer (respectively for the $\left \{  \textit{First}, \textit{Second} \right \}$ networks). $N$ is the number of samples in a batch and $C$ is the total number of classes.

For the morph binary classification component only the definition of the cross label is changed:
\begin{equation}
\label{eq:binary_class}
    L_{3} = -\frac{1}{N} \sum_{i}^{N} t\log \frac{1}{1+e^{-D}} + (1-t)\log \left( 1-\frac{1}{1+e^{-D}} \right),
\end{equation}
where $D = \dot{f} \cdot \ddot{f}$ is a dot product of high level features extracted by \textit{First} and \textit{Second} backbones and the cross-label $t = abs(sgn(\dot{y}2_i-\ddot{y}2_i))$ of the $i-th$ sample pair. 

The above strategy imply pushing morph samples towards their original classes differently by \textit{First} and \textit{Second} networks. This allow to increase the distance between the morph samples in the feature domain.

In this work we also consider a modification of the above approach. We propose to allocate separate classes for the morphs samples. The formulation of such case means redefining the classification labels \textit{for the morph samples} and doubling the number of identity classes $C$:  $\dot{y}1^*_i = \dot{y}1_i + C$; $\ddot{y}2^*_i =\ddot{y}2_i + C$.

Such class allocation is made differently by \textit{First } and \textit{Second} in and does not impact the differentiating morphs by these networks.  However, since it also pushes the morph samples away from their original classes it can help to increase discriminative power of deep face features.
Further in the work we make experiments with both cases and mark the original strategy with the label \textit{V1} and the modified strategy with \textit{V2}

\section{Data Mining} 
\label{Data_Mining}

With the adopted methodology the training can be proceeded on the same data as S-MAD. The only sampling of this data is different. Recall that VGGFace2 dataset \cite{VGGface2}, is used as a source of original bona fide images. We repeat the quality based filtering of this dataset and generate respective morphs.

\subsection{Morph dataset}
In this work we utilize two automatic methods for generating morphs. First we use a customized landmark-based morphing approach with blending coefficient $0.5$. Second we generate GAN-based morphs with use of the StyleGAN\cite{styleGAN} method. To synthesize such morphs for two original images, they are first projected to a latent domain and then their deep representations are interpolated. The resulting morph is generated from such interpolated latent embedding.

To ensure effective learning in the fused classification task (see Fig. \ref{fig:mdf_schematic}), it is crucial to have unambiguous class labeling in our training dataset of morphs, which are generated from face images from different classes. To address this, we follow the original S-MAD approach\cite{MorDeephy} and employ a strategy where the dataset is split into two disjoint parts attributed to the \textit{First} and \textit{Second} networks. When generating face morphs, we randomly pair images from these identity subsets and label the morphed images accordingly for classification by the respective networks. This approach acts as a regularization technique and enhances the performance of morphing detection. By separating the dataset into two disjoint identity sets, we ensure consistent classification of morphed combinations by the networks.

\subsection{Selfmorphing}
Fully automatic landmark morphing methods often introduce visible artifacts to the generated images, which can bias the learning process towards these artifacts. However, real fraudulent morphs are retouched to remove such perceptual artifacts. To address this, we utilize \textit{selfmorphs}, generated by applying face morphing to images of the same identity. We follow the original S-MAD approach\cite{MorDeephy} and use selfmorphs as bona fide samples to focus on the behavior of deep face features rather than detecting artifacts. We assume that the deep discriminative face features remain intact after selfmorphing.

\begin{table*}[h]
\caption{Comparison of Fused Classification (FC) with the Binary Classification (BC) by APCER@BPCER = (0.1, 0.01) in several protocols.}
\setlength\tabcolsep{4pt}
\begin{center}
\begin{tabular}{|c|cccccccc|}
\hline
\multirow{4}{*}{Method} & \multicolumn{8}{c|}{$APCER@BPCER=\delta$}                                                                                                                                                                                   \\ \cline{2-9} 
                        & \multicolumn{2}{c|}{protocol-asml}  
                        & \multicolumn{2}{c|}{protocol-facemorpher}                                  & \multicolumn{2}{c|}{protocol-webmorph}                                  & \multicolumn{2}{c|}{protocol-stylegan}                                               \\ \cline{2-9} 
                        & \multicolumn{1}{c|}{$\delta =$} & \multicolumn{1}{c|}{$\delta =$} & \multicolumn{1}{c|}{$\delta =$} & \multicolumn{1}{c|}{$\delta =$} & \multicolumn{1}{c|}{$\delta =$} & \multicolumn{1}{c|}{$\delta =$} &
                        \multicolumn{1}{c|}{$\delta =$} & {$\delta =$} \\
                        & \multicolumn{1}{c|}{$0.1$} & \multicolumn{1}{c|}{$0.01$} & \multicolumn{1}{c|}{$0.1$} & \multicolumn{1}{c|}{$0.01$} & \multicolumn{1}{c|}{$0.1$} & \multicolumn{1}{c|}{$0.01$} &  \multicolumn{1}{c|}{$0.1$}  & {$0.01$} \\ \hline
   BC                    & \multicolumn{1}{c|}{0.315}      & \multicolumn{1}{c|}{0.729} & \multicolumn{1}{c|}{0.245}      & \multicolumn{1}{c|}{0.649}       & \multicolumn{1}{c|}{0.391}      & \multicolumn{1}{c|}{0.701}       & \multicolumn{1}{c|}{0.913}          &   0.997   \\ \hline
FCV1 & \multicolumn{1}{c|}{0.063}      & \multicolumn{1}{c|}{0.351}& \multicolumn{1}{c|}{0.066}      & \multicolumn{1}{c|}{0.514}       & \multicolumn{1}{c|}{0.135}      & \multicolumn{1}{c|}{0.529}       & \multicolumn{1}{c|}{0.556}       &     0.959   \\ \hline
FCV2 & \multicolumn{1}{c|}{0.039}      & \multicolumn{1}{c|}{0.275} & \multicolumn{1}{c|}{0.061}      & \multicolumn{1}{c|}{0.315}       & \multicolumn{1}{c|}{0.102}      & \multicolumn{1}{c|}{0.4595}       & \multicolumn{1}{c|}{0.501}            &      0.957 \\  \hline
\end{tabular}
\end{center}
\vspace{0mm}
\label{tab:differential_benchmarking}
\end{table*}

\begin{figure*}[!ht]
\centering
    \includegraphics[width=0.99\linewidth]{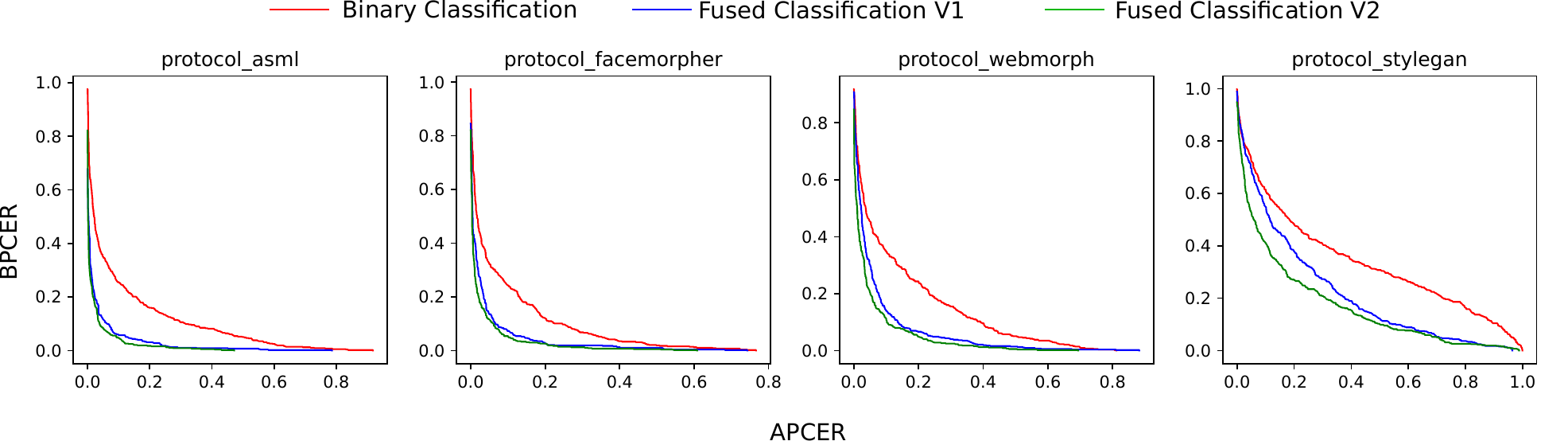} 
    \caption{DET curves for the D-MAD algorithms by NIST FRVT MORPH benchmark. a) protocol Visa-Border; b) protocol Manual; c) protocol MIPGAN-II; d) protocol Print + Scanned.}
\label{fig:result_custom_dmad}
\end{figure*}

In our work \textit{selfmorphs} are generated for both landmark-based and GAN-based morphing approaches.

\subsection{Result dataset}
Our resulting dataset consist of $\sim$500k original images from VGGFace2, $\sim$250k their landmark-based selfmorphs, $\sim$250k their GAN-based selfmorphs, $\sim$500k landmark-based morphs and  $\sim$500k GAN-based morphs. The overall dataset is balanced by the amount of bona fides and morphs. 

\section{Benchmarking.}

One commonly used metric for evaluating single image morphing detection is the relationship between the Bona fide Presentation Classification Error Rate (BPCER) and the Attack Presentation Classification Error Rate (APCER) as specified by ISO/IEC 30107-3 \cite{REFERENCE_OF_ISO_30107}. This relationship can be visualized using a Detection Error Trade-off (DET) curve.

For this work we adopt the public Face Morphing Detection benchmark utilities \cite{MorDeephy}\footnote{https://github.com/iurii-m/MorDeephy} for the differential case. We develop the functionality for generation verification protocols in the differential pipeline and generate several protocols basing on the public data.  Bona fide pairs in all the protocols are combined from the frontal faces of the following public datasets: FRLL Set\cite{Face_Research_Lab_London_Set}, FEI \cite{FEI_dataset}, Aberdeen and Utrecht \cite{Pics_dataset} ($\sim$500 pairs in total).
Morph pairs are combined by pairing images from the morphs from FRLL-Morphs dataset \cite{FRLL_FRGC_Morphs} and bona fides from FRLL Set\cite{Face_Research_Lab_London_Set}. We propose several protocols for different type of morphs (protocol names correspond to the FRLL-Morph subset names): \textit{protocol-asml} ($\sim$ 4.5k morph pairs); \textit{protocol-facemorpher} ($\sim$ 2.5k morph pairs); \textit{protocol-webmorph} ($\sim$ 2.5k morph pairs); \textit{protocol-stylegan} ($\sim$ 2.5k morph pairs).	


Several benchmarks (with restricted data and protocols) available for evaluating the performance of morphing detection or morphing resistant algorithms: The NIST FRVT MORPH \cite{bench_NIST_morph} and FVC-onGoing MAD \cite{bench_1}. They accept both no-reference and differential morphing algorithms, however they are proprietary and managed by a specific entity, leading to submission restrictions and limited accessibility. In this work we will use the public results of NIST FRVT MORPH to compare with our approach.

\begin{table*}[!ht]
\vspace{0mm}
\caption{Comparison with differential image morphing detection methods by APCER@BPCER = (0.1, 0.01) from the NIST FRVT MORPH benchmark.}

\begin{center}
\begin{tabular}{|c|cccccccc|}
\hline
\multirow{3}{*}{Method} & \multicolumn{8}{c|}{$APCER@BPCER=\delta$}                                                                                                                                                                                   \\ \cline{2-9} 
                        & \multicolumn{2}{c|}{Visa-Border}                                  & \multicolumn{2}{c|}{Manual}                                  & \multicolumn{2}{c|}{MIPGAN-II}                                  & \multicolumn{2}{c|}{Print+Scan}             \\ \cline{2-9} 
                         & \multicolumn{1}{c|}{$\delta =$} & \multicolumn{1}{c|}{$\delta =$} & \multicolumn{1}{c|}{$\delta =$} & \multicolumn{1}{c|}{$\delta =$} & \multicolumn{1}{c|}{$\delta =$} & \multicolumn{1}{c|}{$\delta =$} &
                        \multicolumn{1}{c|}{$\delta =$} & {$\delta =$} \\
                        & \multicolumn{1}{c|}{$0.1$} & \multicolumn{1}{c|}{$0.01$} & \multicolumn{1}{c|}{$0.1$} & \multicolumn{1}{c|}{$0.01$} & \multicolumn{1}{c|}{$0.1$} & \multicolumn{1}{c|}{$0.01$} &  \multicolumn{1}{c|}{$0.1$}  & {$0.01$} \\ \hline
Scherhag et al.\cite{face_morphing_MAD}                       & \multicolumn{1}{c|}{0.013 }      & \multicolumn{1}{c|}{0.212}       & \multicolumn{1}{c|}{0.055	}      & \multicolumn{1}{c|}{0.357}       & \multicolumn{1}{c|}{0.004 }      & \multicolumn{1}{c|}{0.134}       & \multicolumn{1}{c|}{0.012}      &    	0.176    \\ \hline
Kashiani et al.\cite{Robust_Ensemble}                        & \multicolumn{1}{c|}{0.447	}      & \multicolumn{1}{c|}{0.901}       & \multicolumn{1}{c|}{0.873	}      & \multicolumn{1}{c|}{0.989}       & \multicolumn{1}{c|}{0.182 }      & \multicolumn{1}{c|}{0.481}       & \multicolumn{1}{c|}{0.842}      &    0.996    \\ \hline
Lorenz et al.\cite{morphing_fusion}                & \multicolumn{1}{c|}{0.432 }      & \multicolumn{1}{c|}{1.000}       & \multicolumn{1}{c|}{0.634 }      & \multicolumn{1}{c|}{1.000}       & \multicolumn{1}{c|}{0.168 }      & \multicolumn{1}{c|}{1.000}       & \multicolumn{1}{c|}{0.732}      &    1.000   \\ \hline
Ferrara et al.\cite{unibo}                    & \multicolumn{1}{c|}{0.966	}      & \multicolumn{1}{c|}{0.999}       & \multicolumn{1}{c|}{0.689	}      & \multicolumn{1}{c|}{0.969}       & \multicolumn{1}{c|}{0.004 }      & \multicolumn{1}{c|}{0.751}       & \multicolumn{1}{c|}{0.070}      &     	0.280   \\ \hline
Ours                      & \multicolumn{1}{c|}{0.232}      & \multicolumn{1}{c|}{0.555}       & \multicolumn{1}{c|}{0.531	}      & \multicolumn{1}{c|}{0.872}       & \multicolumn{1}{c|}{0.359 }      & \multicolumn{1}{c|}{0.859}       & \multicolumn{1}{c|}{0.680}      &  	0.926   \\ \hline
Ours(FR)                       & \multicolumn{1}{c|}{0.087	}      & \multicolumn{1}{c|}{0.453}       & \multicolumn{1}{c|}{-}      & \multicolumn{1}{c|}{-}       & \multicolumn{1}{c|}{-}      & \multicolumn{1}{c|}{-}       & \multicolumn{1}{c|}{0.125}      &  	0.568    \\ \hline
\end{tabular}
\end{center}
\vspace{0mm}
\label{tab:APCER_BPCER_NIST_DMAD}
\end{table*}

\begin{figure*}[!ht]

\centering
    \includegraphics[width=0.95\linewidth]{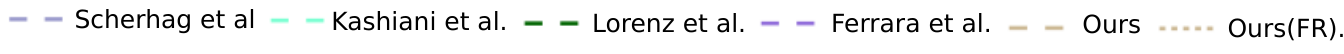}

    \includegraphics[width=0.25\linewidth]{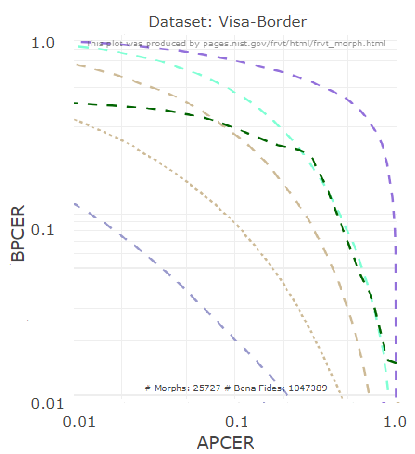}
    \includegraphics[width=0.21\linewidth]{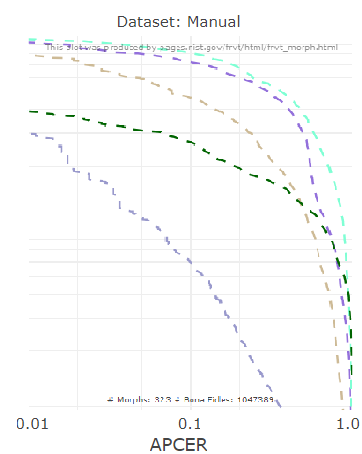}
    \includegraphics[width=0.21\linewidth]{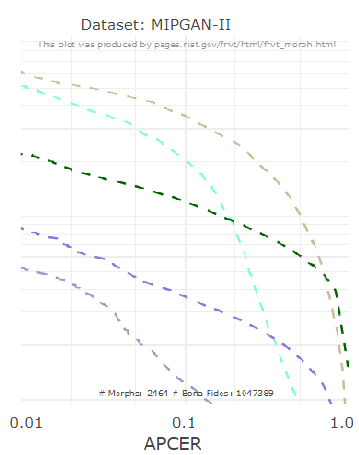}
    \includegraphics[width=0.21\linewidth]{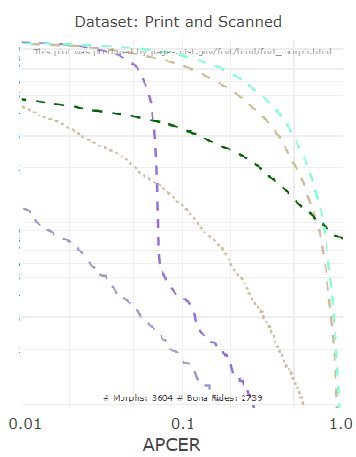}
     
    a \hspace{0.18\linewidth} b \hspace{0.18\linewidth}  c \hspace{0.18\linewidth} d

    \caption{DET curves for the D-MAD algorithms by NIST FRVT MORPH benchmark. a) protocol Visa-Border; b) protocol Manual; c) protocol MIPGAN-II; d) protocol Print + Scanned.}
\label{fig:result_nist_dmad}

\end{figure*}

\section{Experiments} 
\label{Experiments}

\subsection{Differential Benchmarking}
We performed experiments of the fused classification strategy with the binary classification baseline and tested those cases in our custom benchmarks. The baseline is implemented on the same setup (see Fig. \ref{fig:mdf_schematic}) where the identity classification components are disabled and the training is driven by a single loss component in Eq. \ref{eq:binary_class}.

In all the cases we use the EfficientNetB3\cite{EfficientNet} backbone network with input image size 300$\times$300. It is trained with SGD optimizer for 5 epochs with momentum 0.9 and  linearly decreasing learning rate from 0.01 to 0.0001. The batch size is 28.

Our results (see Fig. \ref{fig:result_custom_dmad}, and Table \ref{tab:differential_benchmarking}) demonstrate the superiority of our approach over the baseline. Fused classification allow to generalize the detection performance to the unseen data and scenarios. 
Also we conclude that the V2 strategy (where morphs are disentangled from their original classes) is superior then the V1 and allow to achieve better MAD performance.

\subsection{NIST FRVT MORPH}
We evaluate the performance of our top-performing model (Fused Classification V3) by comparing it with several state-of-the-art (SOTA) D-MAD approaches, which have public results on the FRVT NIST MORPH Benchmark \cite{bench_NIST_morph}.
We perform comparison in several protocols: \textit{Visa-Border} (25727 Morphs);  \textit{Manual} (323 Morphs); \textit{MIPGAN-II} (2464 Morphs); \textit{Print + Scanned} (3604 Morphs).
All protocols in the comparison utilize a substantial collection of $\sim$1M bona fide images. The performance evaluation is conducted using the metrics $APCER@BPCER = (0.1, 0.01)$.

Our performance results (see Table \ref{tab:APCER_BPCER_NIST_DMAD}, Fig. \ref{fig:result_nist_dmad}) are comparable to the leaders in several benchmarks.
Also, our method does not demonstrate bias to a particular morphing generative strategy and has the most stable performance across all protocols in comparison to  other approaches. 

We also present the algorithm (Ours(FR))(see Table \ref{tab:APCER_BPCER_NIST_DMAD}), where the morphing detection signal of our fused classification detector is multiplied with the similarity, which is given by a face recognition model. This algorithm indeed demonstrate a superior result in all the benchmarks, where it was tested.
This indicates that currently differential face morphing benchmarks share many similarities with 1-1 face verification protocols and can be approached with only a strong face recognition model in hands. 
Such approach is reasonable from the practical security perspective of detecting impostors, but not very correct from the academic perspective, since the face morphing detector in differential case is prompted to detect identity non-matched pairs that may not have face morphs at all. 

Despite the fact that this feature does not pose significant risks, it should be taken into account when developing algorithms for differential face morphing detection or benchmarks for their evaluation.







\section{Conclusion} 
\label{Conclusion}

In this paper, we focus on differential face morphing detection and propose a novel deep learning method that incorporates sophisticated face recognition tasks and employs a fused classification scheme for morphs classification. 
We propose public benchmark utilities for differential face morphing detection.
Additionally we and raise several questions on the differences in the vision of the differential face morphing detection in academic and security application perspective.

{\small
\bibliographystyle{ieee_fullname}
\bibliography{egbib}

\begin{thebibliography}{10}\itemsep=-1pt

\bibitem{AttentionMorphing}
P. Aghdaie, B. Chaudhary, S. Soleymani, J. Dawson, and N. Nasrabadi.
\newblock {Attention Aware Wavelet-based Detection of Morphed Face Images}.
\newblock {\em 2021 IEEE IJCB}, pages 1--8, 2021.

\bibitem{FEI_dataset}
{Artificial Intelligence Laboratory of FEI in São Bernardo do Campo, São
  Paulo, Brazil}.
\newblock {FEI face database}, 2006.
\newblock https://fei.edu.br/~cet/facedatabase.html.

\bibitem{ubo_morpher}
{Biometric System Laboratory}.
\newblock {UBO-Morpher}, 2018.
\newblock http://biolab.csr.unibo.it/Research.asp. (accessed: September 1,
  2022).

\bibitem{Double_Siamese_Morphing}
G. Borghi, E. Pancisi, M. Ferrara, and D. Maltoni.
\newblock A double siamese framework for differential morphing attack
  detection.
\newblock {\em Sensors}, 21:3466, 05 2021.

\bibitem{VGGface2}
Q. Cao, L. Shen, W. Xie, O.~M. Parkhi, and A. Zisserman.
\newblock Vggface2: A dataset for recognising faces across pose and age.
\newblock In {\em International Conference on FG}, 2018.

\bibitem{MorDIFF}
N. Damer, M. Fang, P. Siebke, J.~N. Kolf, M. Huber, and F. Boutros.
\newblock Mordiff: Recognition vulnerability and attack detectability of face
  morphing attacks created by diffusion autoencoders, 2023.

\bibitem{morGAN}
N. {Damer}, A.~M. {Saladié}, A. {Braun}, and A. {Kuijper}.
\newblock {MorGAN: Recognition Vulnerability and Attack Detectability of Face
  Morphing Attacks Created by Generative Adversarial Network}.
\newblock In {\em 2018 IEEE 9th International Conference on BTAS}, pages 1--10,
  2018.

\bibitem{PW-MAD}
N. Damer, N. Spiller, M. Fang, F. Boutros, F. Kirchbuchner, and A. Kuijper.
\newblock Pw-mad: Pixel-wise supervision for generalized face morphing attack
  detection.
\newblock In {\em Advances in Visual Computing}, pages 291--304, 2021.

\bibitem{PRNU_2}
L. Debiasi, U. Scherhag, C. Rathgeb, A. Uhl, and C. Busch.
\newblock {PRNU-based detection of morphed face images}.
\newblock {\em 2018 IWBF}, pages 1--7, 2018.

\bibitem{Face_Research_Lab_London_Set}
L. DeBruine and B. Jones.
\newblock Face research lab london set, May 2017.
\newblock {https://figshare.com/articles/dataset/Face\_Research\_Lab\_
  London\_Set/5047666/3.}

\bibitem{arcface_paper}
J. {Deng}, J. {Guo}, N. {Xue}, and S. {Zafeiriou}.
\newblock {ArcFace: Additive Angular Margin Loss for Deep Face Recognition}.
\newblock In {\em 2019 Conference on CVPR}, pages 4685--4694, 2019.

\bibitem{magic_passport}
M. Ferrara, A. Franco, and D. Maltoni.
\newblock {The magic passport}.
\newblock {\em {IJCB 2014 - 2014 IEEE/IAPR}}, 12 2014.

\bibitem{face_demorphing}
M. {Ferrara}, A. {Franco}, and D. {Maltoni}.
\newblock Face demorphing.
\newblock {\em IEEE Transactions on Information Forensics and Security},
  13(4):1008--1017, 2018.

\bibitem{unibo}
M. Ferrara, A. Franco, and D. Maltoni.
\newblock {Face morphing detection in the presence of printing/scanning and
  heterogeneous image sources}.
\newblock {\em IET Biometrics}, 10, 02 2021.

\bibitem{Huang_2020_CVPR}
Y. Huang, Y. Wang, Y. Tai, X. Liu, P. Shen, S. Li, J. Li, and F. Huang.
\newblock {CurricularFace: Adaptive Curriculum Learning Loss for Deep Face
  Recognition}.
\newblock In {\em {The IEEE/CVF Conference on CVPR}}, June 2020.

\bibitem{REFERENCE_OF_ISO_30107}
{International Organization for Standardization}.
\newblock {ISO/IEC 30107–3:2017. Information Technology—Biometric
  Presentation Attack Detection — Part 3: Testing and Reporting}.
\newblock ISO/IEC JTC 1/SC 37 Biometrics, 09 2017.

\bibitem{styleGAN}
T. {Karras}, S. {Laine}, and T. {Aila}.
\newblock {A Style-Based Generator Architecture for Generative Adversarial
  Networks}.
\newblock In {\em 2019 IEEE/CVF Conference on CVPR}, pages 4396--4405, 2019.

\bibitem{Robust_Ensemble}
H. Kashiani, S.~M. Sami, S. Soleymani, and N.~M. Nasrabadi.
\newblock Robust ensemble morph detection with domain generalization.
\newblock In {\em {IEEE} International {IJCB} 2022}, pages 1--10. {IEEE}, 2022.

\bibitem{AdaFace}
M. Kim, A.~K. Jain, and X. Liu.
\newblock Adaface: Quality adaptive margin for face recognition.
\newblock {\em 2022 IEEE/CVF Conference on CVPR}, pages 18729--18738, 2022.

\bibitem{morphing_fusion}
S. Lorenz, U. Scherhag, C. Rathgeb, and C. Busch.
\newblock {Morphing attack detection: A fusion approach}.
\newblock In {\em IEEE Fusion}, 2021.

\bibitem{MorDeephy}
I. Medvedev, F. Shadmand, and N. Gonçalves.
\newblock Mordeephy: Face morphing detection via fused classification.
\newblock In {\em Proceedings of ICPRAM}, pages 193--204. SciTePress, 2023.

\bibitem{QualFace2}
I. Medvedev, J. Tremoço, B. Mano, L.~E. Santo, and N. Gonçalves.
\newblock Towards understanding the character of quality sampling in deep
  learning face recognition.
\newblock {\em IET Biometrics}, 11(5):498--511, 2022.

\bibitem{Magface}
Q. Meng, S. Zhao, Z. Huang, and F. Zhou.
\newblock Magface: A universal representation for face recognition and quality
  assessment.
\newblock In {\em 2021 IEEE/CVF Conference on CVPR}, pages 14220--14229, 2021.

\bibitem{face_morphing_fd}
T. Neubert, C. Kraetzer, and J. Dittmann.
\newblock {A Face Morphing Detection Concept with a Frequency and a Spatial
  Domain Feature Space for Images on eMRTD}.
\newblock In {\em Proceedings of the ACM Workshop}, pages 95--100, 07 2019.

\bibitem{nist_conference_2022}
{NIST}.
\newblock {International Face Performance Conference}, 2022.
\newblock
  https://www.nist.gov/news-events/events/2022/11/international-face-performance-conference-ifpc-2022.
  (accessed: May 1, 2023).

\bibitem{bench_NIST_morph}
{NIST}.
\newblock {NIST FRVT MORPH}, 2022.
\newblock https://pages.nist.gov/frvt/html/frvt\_morph.html.

\bibitem{FMD_Feature_Wise_Supervision}
Le Qin, Fei Peng, and Min Long.
\newblock Face morphing attack detection and localization based on feature-wise
  supervision.
\newblock {\em IEEE TIFS}, 17:3649--3662, 2022.

\bibitem{morphing_spiegel}
Thelen R. and Horchert J.
\newblock {Aktivisten schmuggeln Fotomontage in Reisepass}, 2018.
\newblock
  {https://www.spiegel.de/netzwelt/netzpolitik/biometrie-im-reisepass-peng-kollektiv-schmuggelt-fotomontage-in-ausweis-a-1229418.html}.
  (accessed: May 1, 2023).

\bibitem{detecting_face_morph_1}
R. {Raghavendra}, K.~B. {Raja}, and C. {Busch}.
\newblock {Detecting morphed face images}.
\newblock In {\em {2016 IEEE 8th International Conference on BTAS}}, pages
  1--7, 2016.

\bibitem{face_morphing_dnn}
R. Raghavendra, K.~B. Raja, S. Venkatesh, and C. Busch.
\newblock {Transferable Deep-CNN Features for Detecting Digital and
  Print-Scanned Morphed Face Images}.
\newblock {\em {2017 IEEE Conference on CVPRW}}, pages 1822--1830, 2017.

\bibitem{bench_1}
K. Raja, M. Ferrara, A. Franco, L. Spreeuwers, I. Batskos, F. Wit, M.
  Gomez-Barrero, U. Scherhag, D. Fischer, S. Venkatesh, J.~M. Singh, G. Li, L.
  Bergeron, S. Isadskiy, R. Raghavendra, C. Rathgeb, D. Frings, U. Seidel, F.
  Knopjes, and C. Busch.
\newblock {Morphing Attack Detection - Database, Evaluation Platform and
  Benchmarking}.
\newblock {\em IEEE TIFS}, PP:1--1, 11 2020.

\bibitem{Towards_morphing_detection}
R. Ramachandra, S. Venkatesh, K. Raja, and C. Busch.
\newblock Towards making morphing attack detection robust using hybrid
  scale-space colour texture features.
\newblock In {\em 2019 IEEE 5th International Conference on ISBA}, pages 1--8,
  2019.

\bibitem{ImageNet_cite}
O. Russakovsky, J. Deng, H. Su, J. Krause, S. Satheesh, S. Ma, Z. Huang, A.
  Karpathy, A. Khosla, M. B., A.~C. Berg, and L. Fei-Fei.
\newblock {{ImageNet Large Scale Visual Recognition Challenge}}.
\newblock {\em {IJCV}}, 115(3):211--252, 2015.

\bibitem{FRLL_FRGC_Morphs}
E. Sarkar, P. Korshunov, L. Colbois, and S\'{e}bastien Marcel.
\newblock Vulnerability analysis of face morphing attacks from landmarks and
  generative adversarial networks.
\newblock {\em arXiv preprint}, Oct. 2020.

\bibitem{face_rec_vulnerability_morph}
U. {Scherhag}, R. {Raghavendra}, K.~B. {Raja}, M. {Gomez-Barrero}, C.
  {Rathgeb}, and C. {Busch}.
\newblock {On the vulnerability of face recognition systems towards morphed
  face attacks}.
\newblock In {\em 2017 5th IWBF}, pages 1--6, 2017.

\bibitem{face_morphing_MAD}
U. Scherhag, C. Rathgeb, J. Merkle, and C. Busch.
\newblock {Deep Face Representations for Differential Morphing Attack
  Detection}.
\newblock {\em IEEE TIFS}, 15:3625--3639, 2020.

\bibitem{Pics_dataset}
{School of Natural Sciences University of Stirling}.
\newblock {Psychological Image Collection of Stirling}, 1998.
\newblock http://pics.stir.ac.uk. (accessed: September 1, 2022).

\bibitem{facenet}
F. {Schroff}, D. {Kalenichenko}, and J. {Philbin}.
\newblock {FaceNet: A unified embedding for face recognition and clustering}.
\newblock In {\em 2015 IEEE Conference CVPR}, pages 815--823, 2015.

\bibitem{towards_face_recognition}
Y. Shi, X. Yu, K. Sohn, M. Chandraker, and A. Jain.
\newblock {Towards Universal Representation Learning for Deep Face
  Recognition}.
\newblock In {\em {Proceedings of 2020 IEEE/CVF CVPR}}, pages 6816--6825, 06
  2020.

\bibitem{equalized_margin_paper}
J. {Sun}, W. {Yang}, J. {Xue}, and Q. {Liao}.
\newblock {An Equalized Margin Loss for Face Recognition}.
\newblock {\em {IEEE Transactions on Multimedia}}, pages 1--1, 2020.

\bibitem{deepid2_plus_paper}
Y. Sun, X. Wang, and X. Tang.
\newblock Deeply learned face representations are sparse, selective, and
  robust.
\newblock {\em 2015 IEEE Conference on CVPR}, pages 2892--2900, 2015.

\bibitem{EfficientNet}
Mingxing Tan and Quoc Le.
\newblock {E}fficient{N}et: Rethinking model scaling for convolutional neural
  networks.
\newblock In Kamalika Chaudhuri and Ruslan Salakhutdinov, editors, {\em
  Proceedings of the 36th International Conference on Machine Learning},
  volume~97 of {\em Proceedings of Machine Learning Research}, pages
  6105--6114. PMLR, 09--15 Jun 2019.

\bibitem{iMARS_project}
{The Community Research and Development Information Service}.
\newblock {Image Manipulation Attack Resolving Solutions}, 2020.
\newblock https://cordis.europa.eu/project/id/883356. (accessed: September 1,
  2022).

\bibitem{torkar2023morphing}
Matjaž Torkar.
\newblock Morphing cases in slovenia.
\newblock {NIST IFPS}, 2022.
\newblock Ministry of the Interior Police, Slovenia.

\bibitem{face_morphing_using_general_purpose_fr}
L. {Wandzik}, G. {Kaeding}, and R.~V. {Garcia}.
\newblock {Morphing Detection Using a General- Purpose Face Recognition
  System}.
\newblock In {\em {2018 26th EUSIPCO}}, pages 1012--1016, 2018.

\bibitem{centerface_paper}
Y. Wen, K. Zhang, Z. Li, and Y. Qiao.
\newblock {A Discriminative Feature Learning Approach for Deep Face
  Recognition}.
\newblock In {\em Computer Vision -- ECCV 2016}, pages 499--515, 2016.

\bibitem{MIPGAN_morphing_paper}
H. Zhang, S. Venkatesh, R. Ramachandra, K. Raja, N. Damer, and C. Busch.
\newblock {MIPGAN - Generating Robust and High Quality Morph Attacks Using
  Identity Prior Driven GAN}.
\newblock {\em {ArXiv}}, abs/2009.01729, 2020.

\end{thebibliography}
}

\end{document}